\documentclass[letterpaper]{article} % DO NOT CHANGE THIS
\usepackage[utf8]{inputenc} % Allow UTF-8 Unicode characters
\usepackage{aaai2027}  % DO NOT CHANGE THIS
\usepackage[hyphens]{url}  % DO NOT CHANGE THIS
\usepackage{graphicx} % DO NOT CHANGE THIS
\usepackage{natbib}  % DO NOT CHANGE THIS AND DO NOT ADD ANY OPTIONS TO IT
\usepackage{soul} 
\usepackage{caption} % DO NOT CHANGE THIS AND DO NOT ADD ANY OPTIONS TO IT
\usepackage{algorithm}
\usepackage{algorithmic}
\usepackage{amssymb}
\usepackage{amsfonts}
\usepackage{amsmath}
\usepackage{subcaption}
\usepackage{multirow}

\usepackage{newfloat}
\usepackage{listings}
\DeclareCaptionStyle{ruled}{labelfont=normalfont,labelsep=colon,strut=off} % DO NOT CHANGE THIS
\floatstyle{ruled}
\newfloat{listing}{tb}{lst}{}
\floatname{listing}{Listing}

\usepackage{booktabs}

\newcommand{\method}{StageWAM}
\nocopyright
\title{StageWAM: Joint-Embedding Stage Prediction for World-Action Models in Robot Manipulation}
\author {
    Xiao Liu\textsuperscript{\rm 1}\equalcontrib,
    Yuguang Yang\textsuperscript{\rm 1}\textsuperscript{\rm 2}\equalcontrib,
    Xi Wang\textsuperscript{\rm 3},
    Kai Jiang\textsuperscript{\rm 4},
    Cheng Chi\textsuperscript{\rm 5},
    Yong Xu\textsuperscript{\rm 3},
    Wenchao Ding\textsuperscript{\rm 6},
    Yilun Chen\textsuperscript{\rm 6},
    Yan Wang\textsuperscript{\rm 1}\corresponding
}
\affiliations {
    \textsuperscript{\rm 1}Institute for AI Industry Research (AIR), Tsinghua University\\
    \textsuperscript{\rm 2}School of Electronic Information Engineering, Beihang University\\
    \textsuperscript{\rm 3}AIR Wuxi Innovation Center,
    Tsinghua University\\
    \textsuperscript{\rm 4}School of Artificial Intelligence, Beihang University\\
    \textsuperscript{\rm 5}School of Information, Renmin University of China\\
    \textsuperscript{\rm 6}TARS Robotics
    
}
\begin{document}

\maketitle

\begin{abstract}
Generalist robot policies aim to map multimodal observations and linguistic task instructions to actions across diverse tasks. However, existing methods typically represent the future as a fixed, short video-action chunk. This short-term future captures local scene evolution for action execution, but it does not explicitly describe the stage-level future that specifies how a task should progress from its current stage to the next. We therefore distinguish two complementary futures for robot manipulation: a short-term physical future to capture local scene evolution and a stage-level semantic future to represent task progress. We introduce \textbf{\method{}}, which augments a Motus-based World Action Model (WAM) with \textbf{Stage-JEPA}, a goal-conditioned Joint-Embedding Predictive Architecture (JEPA) predictor. Given the current observation and task instruction, Stage-JEPA uses a frozen V-JEPA2 encoder to extract the current-state representation and predicts the latent target of the next inferred stage. Across 50 RoboTwin 2.0 tasks in clean and randomized environments, \method{} achieves 90.25\% overall success and reduces the mean number of execution steps in successful rollouts by 5.97\% relative to the strongest baseline.
\end{abstract}

\section{Introduction}
\begin{figure*}[t]
\begin{subfigure}[b]{0.29\textwidth}
    \centering
    \includegraphics[width=\linewidth]{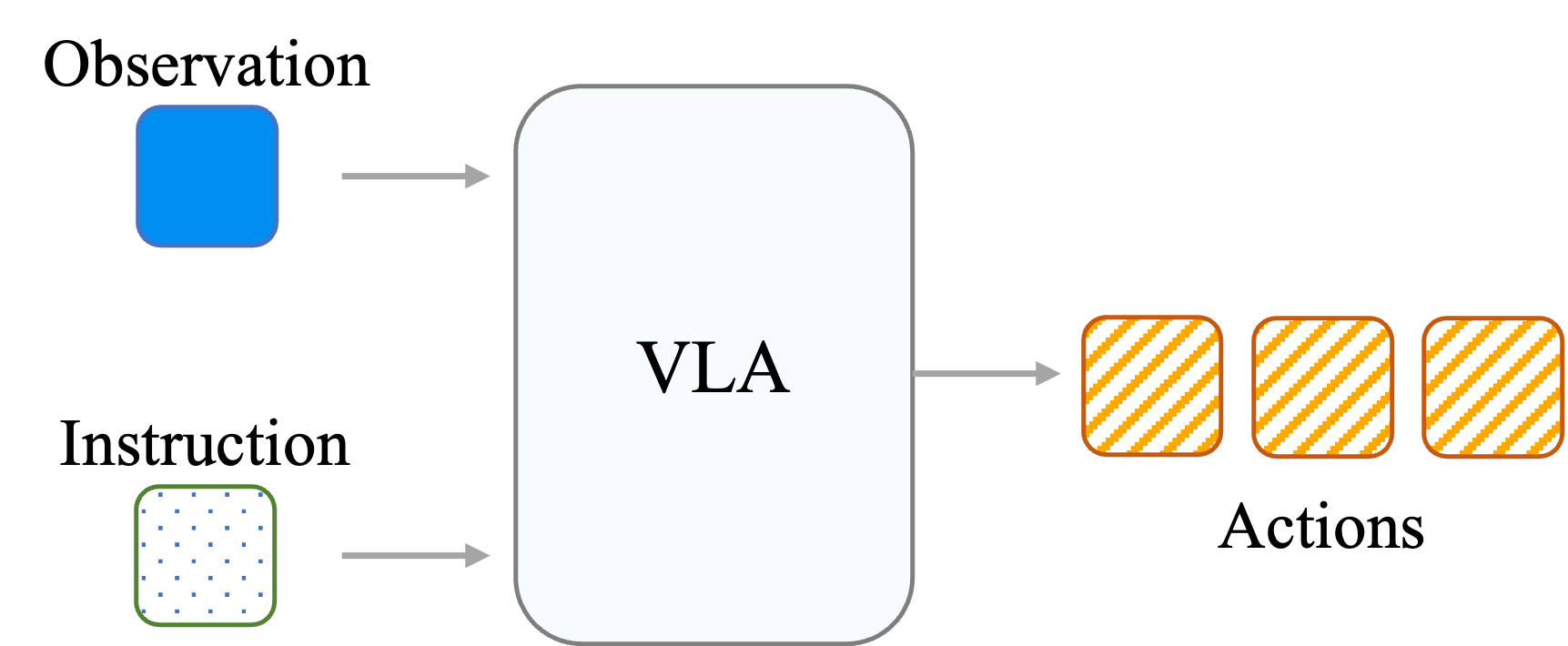}
    \caption{\textbf{VLA.}}
    \label{fig:overview-vla}
\end{subfigure}\hfill
\begin{subfigure}[b]{0.29\textwidth}
    \centering
    \includegraphics[width=\linewidth]{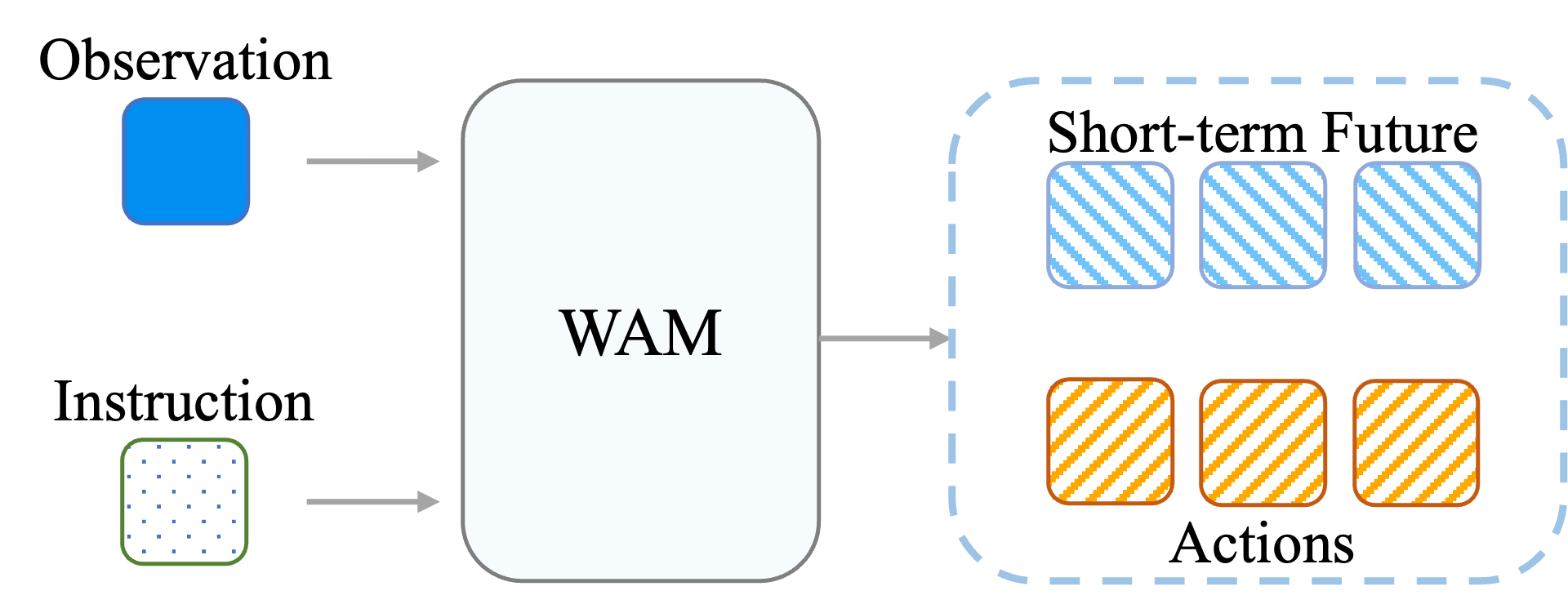}
    \caption{\textbf{WAM.}}
    \label{fig:overview-wam}
\end{subfigure}\hfill
\begin{subfigure}[b]{0.315\textwidth}
    \centering
    \includegraphics[width=\linewidth]{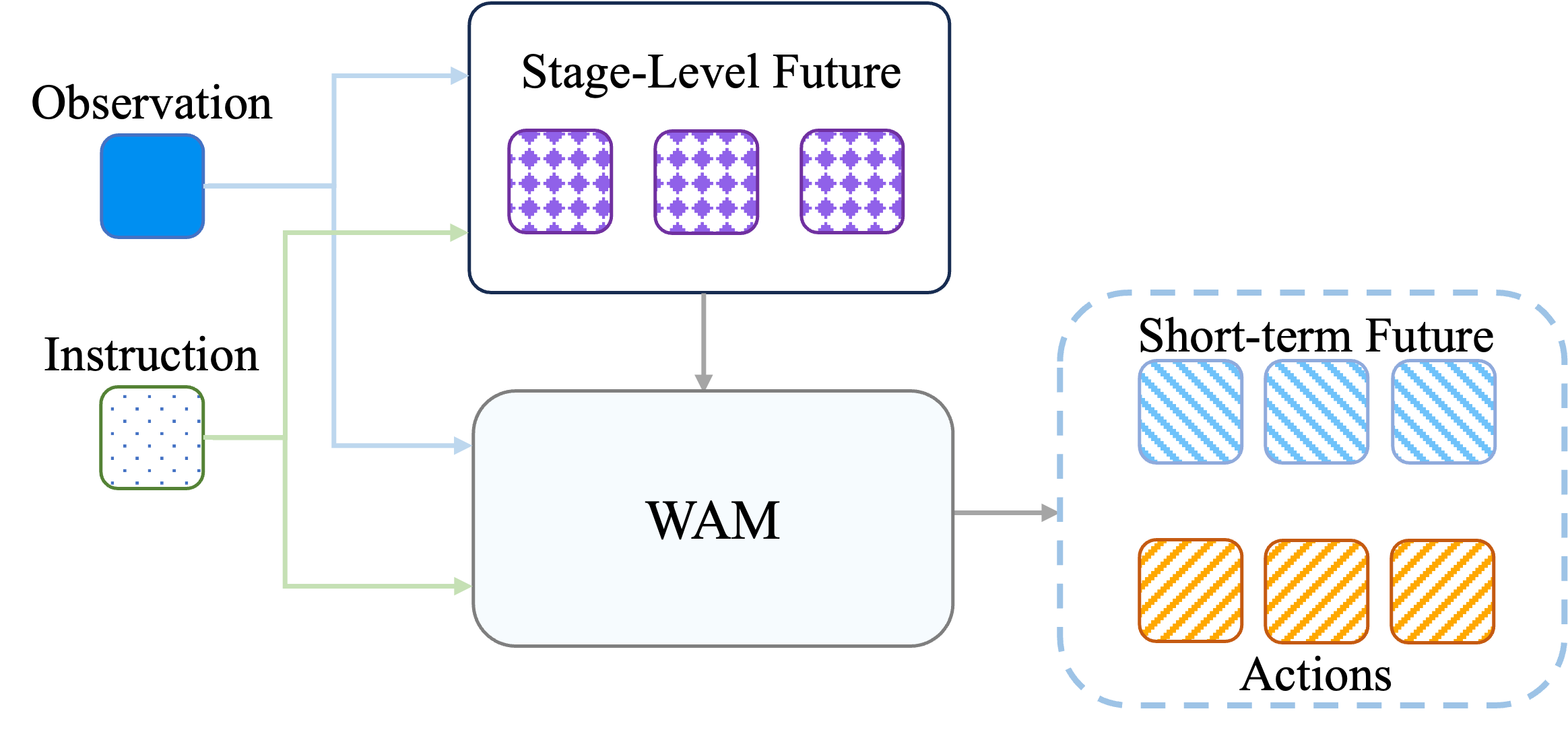}
    \caption{\textbf{\method{}.}}
    \label{fig:overview-jepawam}
\end{subfigure}
\caption{\textbf{Comparison of future modeling in robot policies.}
(a) A vision-language-action policy directly predicts actions from the current observation and task instruction without explicitly modeling future observations.
(b) A world-action model jointly predicts the short-term visual future and the corresponding actions.
(c) \method{} predicts stage guidance from the current observation and task instruction, and uses it as an internal condition for short-term world-action prediction. The stage guidance represents the intended task progress rather than a generated future observation.}
\label{fig:introduction}
\end{figure*}

Generalist robot policies aim to convert visual observations, robot states, and task instructions into executable actions across diverse tasks and environments. Vision-Language-Action models (VLAs)~\citep{zitkovich2023rt2,kim2024openvla,intelligence2025pi_05,zheng2025xvla} transfer semantic representations and instruction-following capabilities from pretrained vision-language models to robot control. However, their direct observation-to-action formulation does not explicitly model how the environment should change as the task progresses. World Action Models (WAMs)\citep{zhu2025unifieduwm,li2025unifieduwa,bi2025motusunifiedlatentaction} address this limitation by jointly predicting future visual observations and robot actions, providing dense supervision for object motion, contact, and local scene evolution. Nevertheless, predicting a plausible video-action trajectory does not mean that the policy understands what state the current task should reach next. 
Existing WAMs typically represent the future as a dense temporal chunk, which can capture local motion and contact dynamics but leaves the next task-relevant object-state or relation change implicit. Figure~\ref{fig:introduction}(a,b) summarizes this contrast: VLAs directly map observations and task instructions to actions, whereas WAMs make the short-term video-action future explicit but still represent task progress through a dense temporal rollout.

This raises a central question: how can a robot policy predict the next meaningful stage of task progress before generating local actions? Such a stage future should encode task-relevant object-state and relation changes, such as spatial alignment or robot--object interaction, without committing to pixel details or transition duration. Directly predicting a distant target image is therefore poorly matched to this goal, since pixel-level prediction can emphasize appearance rather than task completion. We distinguish a \emph{short-term physical future}, which describes immediate video-action evolution, from a \emph{stage-level semantic future}, which specifies the next object or relation state to reach. 

Joint-Embedding Predictive Architectures (JEPAs)~\cite{bardes2023vjepa,assran2025vjepa2} provide a natural mechanism for modeling this stage-level future. Instead of reconstructing every pixel, a JEPA predicts the latent representation of a target state from the current context, preserving task-relevant semantic and physical structure while remaining less sensitive to low-level appearance. Inspired by this, we use a goal-conditioned JEPA predictor to estimate the latent representation of the next task stage inferred from robot demonstrations. As illustrated in Figure~\ref{fig:introduction}c, this predicted latent serves as an internal progress target that conditions short-term world and action generation, rather than as a generated future observation.

\begin{figure*}[t]
\centering
\includegraphics[width=0.9\textwidth]{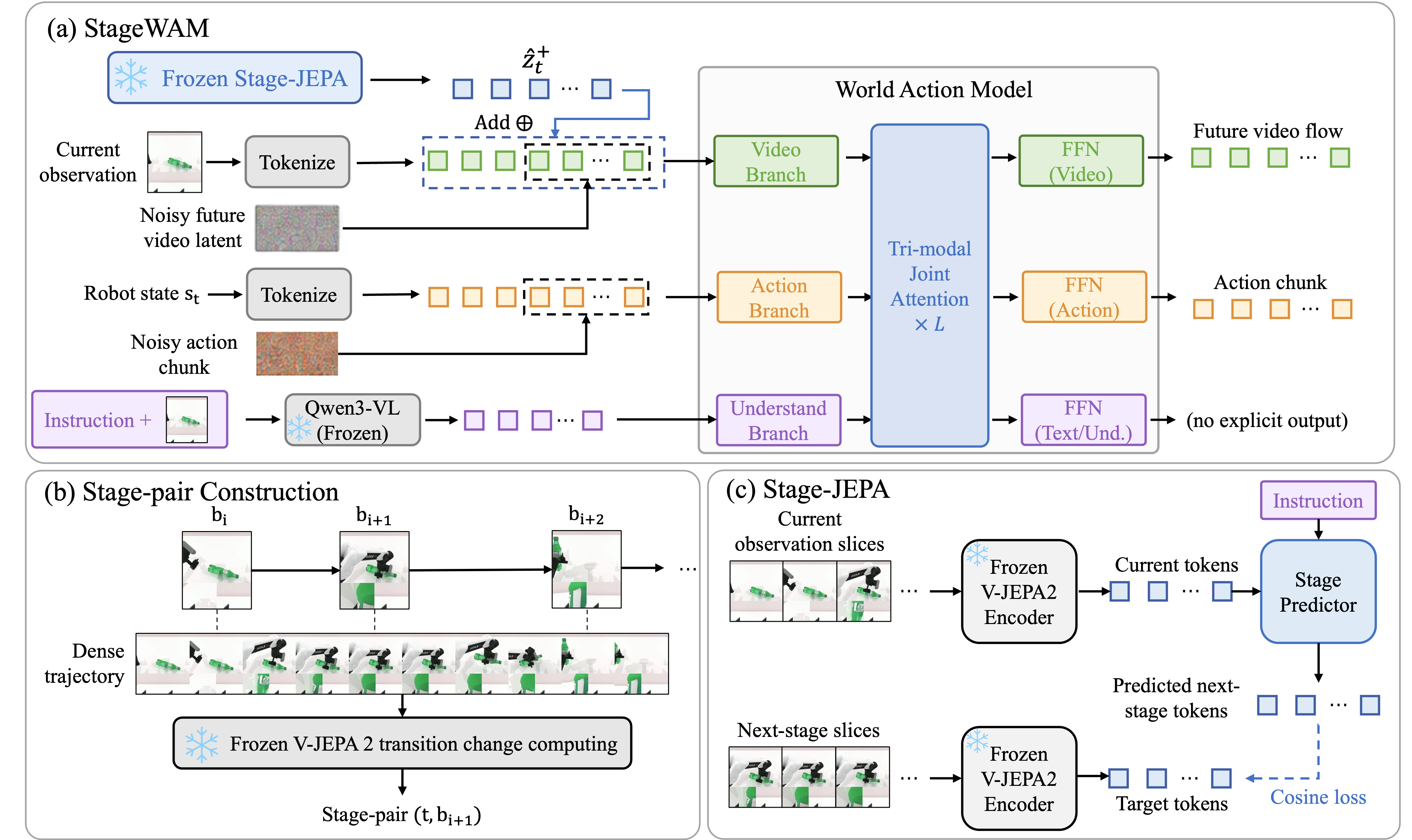}
\caption{\textbf{Overview of \method{}.}
(a) \method{} injects frozen Stage-JEPA guidance into the video stream of a World Action Model, enabling the predicted next-stage latent tokens to condition short-term video-flow and action-chunk prediction.
(b) Stage-pair construction uses frozen V-JEPA2 representation changes to infer stage boundaries from dense robot trajectories and pairs each current frame $t$ with its next boundary $b_{i+1}$.
(c) Stage-JEPA is trained by encoding current and next-stage observation slices with frozen V-JEPA2 and aligning predicted next-stage tokens with target tokens using cosine loss.}
\label{fig:framework}
\end{figure*}

We introduce \method{}, illustrated in Figure~\ref{fig:framework}, which couples a JEPA branch that predicts the next desired task-state transition with a WAM that models the local scene and action dynamics required to realize it. Given the current observation and task instruction, a frozen V-JEPA2 encoder extracts the current representation, and a goal-conditioned JEPA predictor estimates the latent representation of the next progress target. Subsequently, a gated interface injects this predicted stage representation into a Motus-based WAM as a denoising condition, allowing the intended task progress latent to guide both video prediction and action generation. Our contributions can be summarized by:
\begin{itemize}
    \item We formulate robot future modeling at two complementary temporal scales: a short-term physical future for local video-action evolution and a stage-level semantic future for task progress.
    \item We introduce \method{}, in which a goal-conditioned JEPA predictor estimates the next-stage latent and a gated interface conditions a WAM's video and action generation on that prediction.
    \item We conduct an extensive experimental evaluation across 50 RoboTwin tasks under clean and randomized settings, demonstrating the effectiveness of \method{} across diverse semantic manipulation categories.
\end{itemize}

\section{Related Work}

\subsection{Vision-Language-Action Policies}

Recent advances in VLMs~\citep{radford2021learningclip,alayrac2022flamingo,liu2023visualllava,li2023blip,chen2024internvl,wang2024cogvlm,Qwen3-VL} have established a strong foundation for connecting visual observations with language instructions. Building on this foundation, VLAs~\citep{zitkovich2023rt2,kim2024openvla,intelligence2025pi_05,song2026reconvla,zheng2025xvla,kim2026cosmospolicy} extend visual-language understanding to robot control by mapping observations and task instructions directly to executable actions. RT-2~\citep{zitkovich2023rt2} demonstrated how semantic knowledge from large-scale vision-language pretraining can be transferred to robotic control by representing actions as discrete tokens and jointly training on vision-language data and robot trajectories. Octo~\cite{team2024octo} and OpenVLA~\cite{kim2024openvla} further leverage heterogeneous, cross-embodiment robot data to improve transfer across tasks and robotic platforms. More recent approaches, including $\pi_{0.5}$~\citep{intelligence2025pi_05}, ReconVLA~\citep{song2026reconvla}, X-VLA~\cite{zheng2025xvla}, Cosmos Policy~\cite{kim2026cosmospolicy}, and VLA-JEPA~\citep{sun2026vlajepaenhancingvisionlanguageactionmodel}, enhance VLA generalization and execution through large-scale pretraining, spatial representation learning, cross-embodiment adaptation, generative action modeling, and latent future-state prediction. These methods primarily follow a direct observation-to-action formulation, leaving future task states and environmental evolution implicitly encoded within the policy. This limits explicit reasoning about task progress and future goal states. In contrast, \method{} predicts a stage-level semantic future that provides anticipatory guidance for local world modeling and action generation.

\subsection{World Action Models}
World models~\citep{ha2018worldmodels,hafner2019dreamtocontrol,hafner2020masteringatari,hafner2023masteringdiverse} learn how an environment evolves under action or task conditions, providing future information for planning and decision making. UVA~\citep{li2025unifieduwa} and UWM~\citep{zhu2025unifieduwm} jointly model future videos and actions, enabling a single model to support forward dynamics, inverse dynamics, and policy learning. Motus~\cite{bi2025motusunifiedlatentaction} unifies visual-language understanding, video generation, and action prediction through multi-expert interaction, while MotuBrain~\cite{team2026motubrain} extends this framework toward cross-embodiment learning and efficient closed-loop deployment. Recent variants improve efficiency or robustness with latent futures, 4D geometric priors, object-addressable slots, or adaptive predictive/reactive control~\citep{chen2026lawamlatentworldaction,zhang2026learning4dgeometricpriors,liu2026oawamobjectaddressableworldaction,feng2026harmowamharmonizinggeneralizableprecise}. These methods mainly refine local dynamics or object-level representations within the WAM horizon, whereas \method{} supplies a predicted next-stage JEPA latent as a task-progress target for WAM generation.

\subsection{Hierarchical and Stage-Guided Manipulation}

Recent work has introduced explicit intermediate goals to improve long-horizon manipulation. Subgoal Diffuser generates coarse-to-fine subgoals to guide model-predictive control, while TaKSIE uses task-progress knowledge to generate visual subgoals for manipulation policies~\cite{huang2024subgoaldiffuser,kang2025incorporatingtaskprogress}. VISTA~\citep{long2026scalingworldmodelforhierarchical} uses a pretrained world model to decompose a task into a sequence of visual subgoals that guide a low-level VLA, and WorldDP~\citep{goswami2026unifyingobjectcentricworldmodels} uses a high-level object-centric world model to optimize feasible subgoals for a low-level diffusion policy. DexFuture~\citep{blark2026dexfuture} predicts structured future visuomotor target trajectories for target-conditioned dexterous control, whereas H-WM~\citep{huang2026h-wm} jointly predicts logical and visual state transitions to provide intermediate guidance to VLA policies. StaKe~\citep{xu2026improvingvisionlanguageactionmodelfinetuning} introduces structured stage and keyframe supervision for long-horizon manipulation.These methods demonstrate the value of modeling task progress beyond a short action horizon, but they typically use explicit visual subgoals, optimized subgoals, or reference trajectories to guide a separate low-level controller. In contrast, \method{} uses joint-embedding prediction to infer a latent target for the next task stage from automatically inferred stage boundaries, and injects this target directly into a WAM to jointly guide local video prediction and action generation.

\section{Task Formulation}
\label{sec:task}

\subsection{Language-Conditioned Manipulation}
Consider a language-conditioned manipulation trajectory $\tau=(o_0,s_0,a_0,\ldots,o_{T-1},s_{T-1},a_{T-1})$, where $o_t$ is the visual observation, $s_t\in\mathbb{R}^{d_s}$ is the robot state, $a_t\in\mathbb{R}^{d_a}$ is the action, $d_s$ and $d_a$ are the state and action dimensions and $l$ is the linguistic task instruction. Classical behavior-cloning policies map the current inputs to an $H_a$-step action chunk,
\begin{equation}
 A_t=(a_t,\ldots,a_{t+H_a-1})=\pi(o_t,s_t,l).
\end{equation}
World--action models additionally predict the short-term visual future $V_t^{\mathrm{short}}=(o_{t+1},\ldots,o_{t+H_v})$ together with this action chunk. A WAM parameterized by $\phi$ therefore models
\begin{equation}
 p_\phi(V_t^{\mathrm{short}},A_t\mid o_t,s_t,l),
 \label{eq:wam-formulation}
\end{equation}
where $H_v$ is the visual horizon. At deployment, the policy executes a finite action chunk, observes the updated state, and replans in a receding-horizon loop. 

\subsection{Stage-Conditioned Manipulation}
The local future $V_t^{\mathrm{short}}$ describes how the scene evolves over a fixed horizon, but it does not explicitly specify the next meaningful task state. We introduce a complementary \emph{stage-level future} $Y_t^{\mathrm{stage}}$ that represents the next task-progress target. A future prediction model $f_\theta$ predicts this target from the current observation and task goal:
\begin{equation}
 \widehat Y_t^{\mathrm{stage}}=f_\theta(o_t,l).
 \label{eq:generic-stage-future}
\end{equation}
We condition the WAM on this predicted target:
\begin{equation}
 p_\phi(V_t^{\mathrm{short}},A_t
 \mid o_t,s_t,l,\widehat Y_t^{\mathrm{stage}}).
 \label{eq:stage-conditioned-wam}
\end{equation}
Thus, the stage-level future specifies \emph{what} task-progress target should be reached, while the short-term visual future and action chunk specify \emph{how} to realize it locally.

\section{Method}

\method{} has two sequential training stages. Stage I trains a goal-conditioned JEPA predictor to infer the next-stage future from the current observation and task goal. Stage II freezes this predictor and trains a WAM to generate the short-horizon visual future and action chunk conditioned on the predicted stage. At inference, the two components run sequentially in a receding-horizon loop.

\subsection{Stage I: Training the JEPA Predictor}
\label{sec:stage-pairs}
Stage-level prediction provides guidance about task evolution rather than its detailed realization. It captures semantic changes in object states and relations, leaving local appearance and motion to the WAM. We therefore learn this guidance with representation-space supervision before introducing the visual and action objectives of Stage II.

\paragraph{Stage-pair construction.}
Stage I constructs supervision from robot trajectories without manual stage labels. To detect key stages, we use a simple JEPA-feature-based transition heuristic to automatically construct stage supervision. For an episode, we sample candidate centers $\mathcal C=\{c_j\}_{j=1}^{M}$ every eight frames and encode a centered 64-frame clip at each center with frozen V-JEPA2, yielding token representations $U_j=\{u_{j,n}\}_n$. We obtain the clip-level representation $v_j$ by pooling these tokens. We score representation change across a local window of $w=2$ candidate positions, where $j^-=\max(1,j-w)$ and $j^+=\min(M,j+w)$:

\begin{align}
 q_j=\max\!\Bigg\{&\mathcal N\!\left(\lVert v_{j^+}-v_{j^-}\rVert_2\right),\\
 &\mathcal N\!\left(\max_n\lVert u_{j^+,n}-u_{j^-,n}\rVert_2\right)\Bigg\},
 \label{eq:stage-transition-scores}
\end{align}
where $\mathcal N$ is per-episode min--max normalization. Temporal NMS with radius $r=3$ retains at most $K=5$ transition keyframes:
\begin{equation}
 \mathcal K=\{c_j\mid j\in\operatorname{NMS}(\{q_j\}_{j=1}^{M};r,K)\}.
 \label{eq:temporal-nms}
\end{equation}
Together with the initial and terminal frames, these keyframes define $\mathcal B=(b_0,\ldots,b_L)=\operatorname{sort}(\{0\}\cup\mathcal K\cup\{T-1\})$. For every stage $[b_i,b_{i+1})$, each current frame $t$ is paired with its next boundary:
\begin{equation}
 \mathcal P=\bigcup_{i=0}^{L-1}\{(t,b_{i+1})\mid b_i\le t<b_{i+1}\}.
 \label{eq:dense-stage-pairs}
\end{equation}
Thus, the keyframes only define boundaries; Stage I is trained with dense current-to-next-stage pairs $(t,b_{i+1})$. The supplementary appendix reports diagnostic statistics of stage-boundary distances and the overlap induced by centered 64-frame training slices.

\paragraph{Goal-conditioned stage prediction.}
\method{} realizes the future prediction model $f_\theta$ as Stage-JEPA, which uses a frozen V-JEPA2 encoder and a trainable goal-conditioned predictor initialized from the V-JEPA2 predictor. Since V-JEPA2 requires 64-frame inputs, a dense pair $(t,b_{i+1})$ is converted into a 64-frame current observation slice $\mathcal S_t$ centered at $t$ and a 64-frame target observation slice $\mathcal S_t^{\mathrm{target}}$ centered at $b_{i+1}$. The same frozen encoder $E$ independently produces the current latent $z_t=E(\mathcal S_t)$ and target latent $z_t^{\mathrm{target}}=E(\mathcal S_t^{\mathrm{target}})$. The predictor produces token-level next-stage predictions $\widehat Z_t^{\mathrm{target}}$, which are pooled and normalized as $\widehat z_t^{\mathrm{target}}$ for Stage-I supervision. Only the predictor and instruction adapter are optimized. For a batch of $B$ normalized prediction-target pairs $(\widehat z_i^{\mathrm{target}},z_i^{\mathrm{target}})$, we optimize
\begin{align}
 \mathcal L_{\mathrm{stage}}
 =\frac{1}{B}\sum_{i=1}^{B}\left(1-(\widehat z_i^{\mathrm{target}})^\top z_i^{\mathrm{target}}\right).
\end{align}
Here, $z_i^{\mathrm{target}}$ is the paired next-stage target for $\widehat z_i^{\mathrm{target}}$, and $\mathcal L_{\mathrm{stage}}$ directly aligns the predicted latent with this target in representation space. The predicted latent realizes the stage target by applying $f_\theta$ to the slice $\mathcal S_t$ and instruction $l$.

\subsection{Stage II: Training the Stage-Conditioned WAM}
The WAM complements stage guidance with the fine-grained visual dynamics and actions needed for local execution. Because joint optimization could let these detail-oriented losses alter the semantic target, we freeze JEPA and train only the WAM and conditioning interface. The current observation slice and instruction produce $\widehat z_t^{\mathrm{target}}\equiv\widehat Y_t^{\mathrm{stage}}$.

We instantiate the WAM with Motus~\citep{bi2025motusunifiedlatentaction}, a Mixture-of-Transformers~\citep{liang2024mixture} architecture comprising a video-generation expert, an action expert, and a vision-language understanding expert. The video expert is initialized from Wan2.2~\citep{wan2025} and models local visual evolution; the action expert represents the robot state and a dense action chunk; and the understanding expert extracts instruction- and observation-dependent features with a frozen Qwen3-VL~\cite{Qwen3-VL} encoder. At every transformer layer, Tri-modal Joint Attention allows the three streams to exchange information, while modality-specific normalization, feed-forward layers, and output heads preserve their specialized functions. This organization is important for control because language and scene understanding can inform both the anticipated visual change and the action sequence, while the video expert supplies fine-grained physical and appearance cues that are absent from a purely semantic policy representation.

During training, the Wan2.2 VAE encodes the current observation and target future observations into video latents. Noise is applied separately to the future-video latent and the demonstrated action chunk, while the current observation, robot state, and instruction remain as conditions. The resulting video, action, and understanding streams are processed jointly, and their respective heads predict the flow fields for recovering the local future video and action chunk. The separate video and action streams retain modality-appropriate representations and objectives, whereas their repeated interaction through joint attention couples predicted scene evolution to executable control.

\method{} augments this WAM before its transformer layers. The video branch forms current-conditioned noisy video tokens $X_t^v\in\mathbb R^{N_v\times d_v}$, and the Stage-JEPA output $\widehat Z_t^{\mathrm{target}}\in\mathbb R^{N_j\times d_j}$ contains $N_j$ predicted stage positions of dimension $d_j$. We mean-pool this output and use an MLP $P_\psi$ to match the Wan video-token dimension before additive injection:
\begin{align}
 h_t &= P_\psi\!\left(\operatorname{Pool}(\widehat Z_t^{\mathrm{target}})\right)
 \in\mathbb R^{d_v},\\
  \alpha &=0.2\sigma(\beta),\\
 \widetilde X_t^v &= X_t^v+\alpha h_t.
\end{align}
Here, $\beta$ is a zero-dimensional trainable parameter and $\alpha$ is a global bounded scalar gate shared by all samples, layers, video tokens, and channels. The condition vector $h_t$ remains sample-specific and is broadcast to all video tokens of that sample. The downstream WAM then predicts the local visual future and action chunk. The same interface applies to other WAMs with an accessible visual conditioning stream.

Training retains the WAM's native visual and action objectives and optionally includes an implementation-specific regularizer for the conditioning interface:
\begin{align}
\mathcal L&= \mathcal L_{\mathrm{video}}+
 \mathcal L_{\mathrm{action}}+\mathcal L_{\mathrm{reg}},\\
 &=\lambda_g \alpha^2+\lambda_\rho(\rho-\rho_0)^2, 
\rho=\frac{\lVert \alpha h_t\rVert}{\lVert X_t^v\rVert}.
\end{align}
Here, $\rho$ measures the magnitude of the injected stage feature relative to the original video tokens, and $\rho_0$ is a manually selected target ratio. The first term biases the bounded gate toward a small residual update, whereas the second discourages the relative update from departing from $\rho_0$.

The two stages thus learn complementary information: Stage I specifies \emph{what task-level change should occur}, and Stage II learns \emph{how to realize it} through detailed visual evolution and actions.

\subsection{Closed-Loop Inference}
At deployment, the policy maintains a causal 64-frame observation buffer $\mathcal H_t$, as required by V-JEPA2. Before 64 observations are available, the earliest observed frame is repeated to pad the temporal slice; no future observations are used. At each policy query, it performs
\begin{align}
 \widehat z_t^{\mathrm{target}}
&= f_\theta(\mathcal H_t,l),\\
(\widehat V_t^{\mathrm{short}},\widehat A_t)
&=W_\phi(o_t,s_t,l,\widehat z_t^{\mathrm{target}}).
\end{align}
Here, $f_\theta$ is the frozen JEPA branch learned in Stage I, and $W_\phi$ is the deployed predictor corresponding to the conditional WAM distribution in Equation~\ref{eq:stage-conditioned-wam}. The predicted stage latent passes through the same conditioning interface as in Stage II, after which the WAM predicts the finite-horizon visual future and action chunk. The environment executes the action chunk, appends the resulting observations to the buffer, and queries the policy again. Stage boundaries and target observation slices are therefore training-time supervision only; inference repeatedly predicts its own stage condition from observed history.

\section{Experiment}
\subsection{Datasets and Baselines}

\paragraph{Datasets.}
We use RoboTwin~\citep{chen2025robotwin}, a dual-arm manipulation benchmark with diverse generated environments. Our study covers 50 tasks under clean and randomized configurations. For evaluation, we run 100 closed-loop episodes for every task--configuration pair with early stopping disabled, yielding 5,000 episodes per configuration and 10,000 episodes per checkpoint. The primary metric is task success rate.

\paragraph{Baselines.}
On RoboTwin, we compare with GO-1~\citep{bu2025agibotgo1}, an AgiBot embodied foundation model based on a vision-language-latent-action formulation; $\pi_{0.5}$~\citep{intelligence2025pi_05}, a generalist VLA that directly predicts action chunks from visual observations and language instructions; X-VLA~\citep{zheng2025xvla}, which strengthens VLA control through large-scale cross-task and cross-embodiment pretraining; and Motus~\citep{bi2025motusunifiedlatentaction}, a WAM that jointly models language-conditioned understanding, future video, and action prediction. The baseline numbers are published results in~\citep{bi2025motusunifiedlatentaction}, and \method{} and Motus are evaluated with the same closed-loop protocol.

\subsection{Implementation Details}

All experiments use eight NVIDIA A800 80GB GPUs with an effective batch size of 256 in both stages. Stage-JEPA uses \texttt{facebook/vjepa2-vitl-fpc64-256} as the stage representation model: the V-JEPA2 encoder is frozen and executed in FP16, while the learnable modules, including the pretrained V-JEPA2 predictor and the task-instruction cross-attention adapter, are trained in FP32 for 1,000 optimization steps with learning rate $1\times10^{-5}$ and weight decay 0.01. We then condition the Motus policy on the trained Stage-JEPA output. The local action policy takes 8-frame videos at resolution $384\times320$, predicts eight future video frames and 16 actions, and injects the stage condition into video tokens through a global scalar gate $\alpha=0.2\sigma(\beta)$ initialized to 0.02. It is trained in BF16 for four epochs with learning rate $1\times10^{-6}$, weight decay 0.01, and conditioning regularizer weights $\lambda_g=10^{-4}$, $\lambda_\rho=10^{-3}$, and $\rho_0=0.08$. Additional gate, token-sampling, and model-selection details are provided in the supplementary appendix.

\subsection{Main Results}

\begin{table*}[t]
\centering
\begingroup
\small
\setlength{\tabcolsep}{0.5mm}
\begin{tabular}{@{}l*{10}{r}@{}}
\toprule
\textbf{Dominant skill (\# tasks)}
  & \multicolumn{2}{c}{\textbf{GO-1}}
  & \multicolumn{2}{c}{\boldmath$\pi_{0.5}$}
  & \multicolumn{2}{c}{\textbf{X-VLA}}
  & \multicolumn{2}{c}{\textbf{Motus}}
  & \multicolumn{2}{c}{\textbf{\method{}}} \\
\cmidrule(lr){2-3}
\cmidrule(lr){4-5}
\cmidrule(lr){6-7}
\cmidrule(lr){8-9}
\cmidrule(l){10-11}
  & Clean & Rand. & Clean & Rand. & Clean & Rand. & Clean & Rand. & Clean & Rand. \\
\midrule
Acquisition \& lifting (5) & 76.40 & 76.60 & 36.80 & 36.20 & 80.80 & 74.20 & \textbf{94.20} & \textbf{94.60} & 92.60 & 92.00 \\
Handover (2) & 10.50 & 10.00 & 23.00 & 18.50 & 36.50 & 18.50 & 82.00 & 68.00 & \textbf{96.50} & \textbf{93.50} \\
Targeted placement (13) & 25.00 & 23.23 & 42.46 & 46.00 & 78.85 & 79.77 & 82.62 & 85.08 & \textbf{88.23} & \textbf{86.54} \\
Container packing (8) & 36.88 & 40.25 & 34.75 & 37.00 & 74.38 & 73.25 & 88.62 & 87.25 & \textbf{91.12} & \textbf{90.00} \\
Arrangement \& stacking (8) & 17.62 & 17.25 & 43.62 & 42.38 & 63.00 & 64.75 & 90.12 & 88.00 & \textbf{91.88} & \textbf{89.62} \\
Articulated/device interaction (7) & 61.43 & 54.71 & 47.86 & 47.43 & 78.43 & 82.43 & \textbf{93.57} & 89.86 & 90.71 & \textbf{90.00} \\
Tool use \& dynamic manipulation (7) & 42.71 & 37.71 & 57.86 & 58.43 & 70.57 & 73.71 & 91.29 & 86.43 & \textbf{95.57} & \textbf{87.86} \\
\midrule
\textbf{All tasks (50)} & 37.86 & 36.24 & 42.98 & 43.84 & 72.88 & 72.84 & 88.66 & 87.02 & \textbf{91.42} & \textbf{89.08} \\
\bottomrule
\end{tabular}
\endgroup
\caption{RoboTwin 2.0 success rates (\%) grouped by the dominant semantic manipulation skill. The number of tasks is shown in parentheses. Categories form a mutually exclusive partition of all 50 tasks, and each entry is the unweighted mean over its tasks. Baseline per-task results are reproduced from Motus~\cite{bi2025motusunifiedlatentaction}; complete task-level results are provided in the supplementary appendix. Bold denotes the highest rate in each row and setting.}
\label{tab:main}
\end{table*}

To examine performance across different manipulation semantics, we group the 50 RoboTwin tasks by their linguistic task instructions into seven categories: acquisition and lifting, handover, targeted placement, container packing, arrangement and stacking, articulated/device interaction, and tool use and dynamic manipulation. Table~\ref{tab:main} reports the mean success rate within each category. Complete task-level results are provided in the supplementary appendix.

In comparison, \method{} achieves 91.42\% success in clean settings and 89.08\% in randomized settings, outperforming Motus in five of seven semantic categories under the clean setting and six under randomization. The strongest results occur on tasks whose execution depends on an explicit change in task progress. Handover, targeted placement, and container packing require the policy to maintain a desired object ownership or object-target relation across transfer and alignment. Arrangement and stacking require ordered intermediate relations, while tool use and dynamic manipulation require recognizing whether the robot should approach, align, or interact. These tasks directly match the two-scale design of \method{}: Stage-JEPA specifies \emph{what task-relevant state should be reached next}, and the WAM models \emph{how to realize it} through local visual dynamics and actions. In contrast, acquisition and lifting depend primarily on immediate grasp geometry and contact stability, leaving less complementary stage structure for JEPA conditioning. Articulated/device interaction contains meaningful state changes, but success also requires mechanism-specific contact, constrained motion, and precise timing, for which the current indirectly conditioned stage signal can be insufficient. Taken together, the improvements across diverse semantic task families indicate that stage-level future guidance is a useful complement to short-horizon WAM prediction.

\subsection{Ablation Study}

\begin{table}[t]
\centering
\footnotesize
\setlength{\tabcolsep}{3pt}
\begin{tabular}{@{}p{0.45\linewidth}ccc@{}}
\toprule
\textbf{Configuration} & \textbf{Clean} & \textbf{Randomized} & \textbf{Overall} \\
\midrule
w/o JEPA & 87.80 & 86.66 & 87.23 \\
future guidance on current-frame tokens only & 91.10 & 88.68 & 89.89 \\
w/o Stage-I training & 91.34 & 88.62 & 89.98 \\
w/o gate regularizers & 91.00 & 88.76 & 89.88 \\
\textbf{\method{} (ours)} & \textbf{91.42} & \textbf{89.08} & \textbf{90.25} \\
\bottomrule
\end{tabular}
\caption{Ablation of \method{} on RoboTwin 2.0. Success rates (\%) are averaged over 50 tasks. Bold denotes the best reported result.}
\label{tab:ablation}
\end{table}

Table~\ref{tab:ablation} isolates the main components of \method{} through four ablations:
\begin{itemize}
    \item \emph{w/o JEPA} removes the JEPA predictor and its stage-guidance input entirely;
    \item \emph{w/o future-latent injection} keeps the Stage-JEPA branch but does not inject the predicted next-stage latent into the WAM video tokens, using the signal only in the current-observation representation;
    \item \emph{w/o Stage-I training} keeps the same conditioning architecture and V-JEPA2 initialization but does not optimize the JEPA predictor on current-to-next-stage pairs;
    \item \emph{w/o gate regularizers} removes the two auxiliary gate losses, $\lambda_g \alpha^2$ and $\lambda_\rho(\rho-\rho_0)^2$, while leaving the gate architecture unchanged.

\end{itemize} 

Removing JEPA produces the largest degradation, indicating that the additional stage-conditioning pathway provides information beyond the short-horizon WAM backbone. Keeping the JEPA branch but preventing future-latent injection also lowers success, which supports injecting the predicted next-stage latent into the WAM representation rather than using it only to modify the current observation. The model without Stage-I training remains competitive, suggesting that the pretrained V-JEPA2 predictor already supplies useful generic visual structure; nevertheless, training on current-to-next-stage pairs gives the best clean, randomized, and overall results. Removing both gate regularizers produces a smaller drop, indicating that the sigmoid-bounded gate already stabilizes the residual update while the two regularization terms provide additional constraints on its scale. Overall, the ablations support both the presence of a stage-conditioning pathway and its direct integration into the WAM video stream, while indicating that the task-specific Stage-I objective provides an additional but comparatively smaller gain.

\begin{figure*}[t]
    \centering
    \begin{subfigure}[t]{0.9\linewidth}
        \centering
        \includegraphics[width=\linewidth]{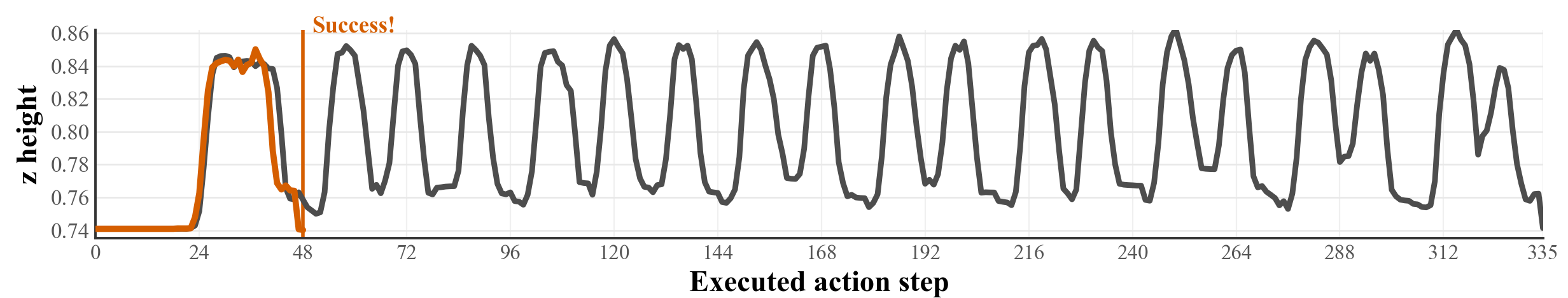}
        \caption{Execution trace of the manipulated object's $z$ height, where orange denotes \method{} and gray denotes Motus. \method{} reaches success much earlier, while Motus exhibits repeated vertical oscillations before eventual success.}
        \label{fig:case-study-trace}
    \end{subfigure}

    \vspace{0.5em}

    \begin{subfigure}[t]{0.9\linewidth}
        \centering
        \includegraphics[width=\linewidth]{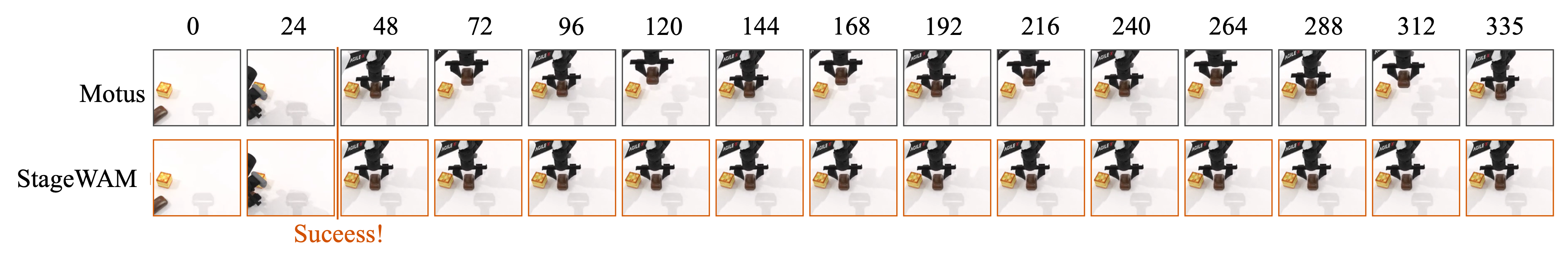}
        \caption{Visual rollout comparison. \method{} quickly advances to the task-completion stage under stage-level guidance, whereas Motus performs repeated local corrections. Frames after \method{} succeeds are padded with the terminal frame for visualization.}
        \label{fig:case-study-rollout}
    \end{subfigure}

    \caption{\textbf{Qualitative case study of stage-guided execution.}
    \method{} uses the predicted next-stage latent to guide local action generation and reaches the success condition with fewer executed steps. In contrast, Motus eventually succeeds only after a longer trajectory with repeated up-and-down motion of the manipulated object.}
    \label{fig:case-study}
\end{figure*}

\subsection{Trajectory analysis.}
Figure~\ref{fig:case-study} provides a qualitative comparison between \method{} and Motus on a representative successful rollout. Although both methods eventually satisfy the success condition, their execution patterns differ substantially. In the execution trace, \method{} reaches success after a short sequence of actions, while Motus continues for a much longer horizon and repeatedly moves the manipulated object up and down before completing the task. The visual rollout shows the same pattern: \method{} quickly advances from the initial interaction to the next task-relevant state, whereas Motus performs several local corrections around the object before reaching the terminal state. This example illustrates the role of the predicted stage latent as an internal progress target. Instead of relying only on short-horizon motion prediction, the WAM receives a representation of the next meaningful stage, which can bias local actions toward task progress and reduce redundant corrective motions.

\subsection{Execution Efficiency}

\begin{figure}[t]
    \centering
    \includegraphics[width=0.9\columnwidth]{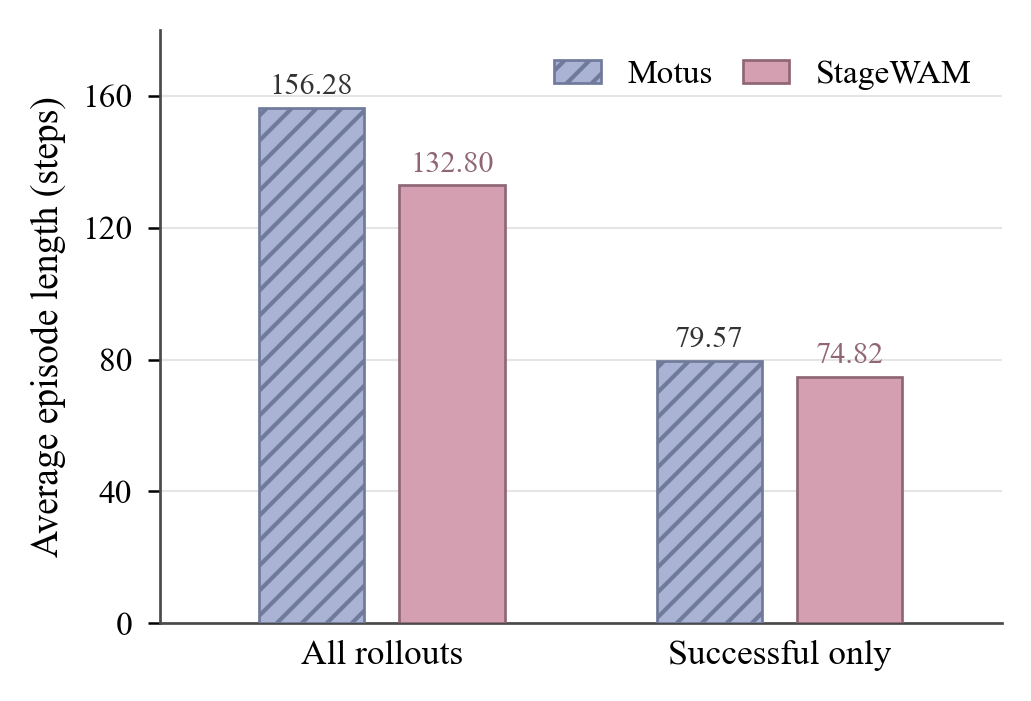}
    \caption{Execution efficiency on RoboTwin 2.0. We report the average episode length over all evaluation rollouts and over successful rollouts only; lower is better.}
    \label{fig:execution-efficiency}
\end{figure}

Beyond task success, we evaluate whether stage guidance enables more efficient execution by measuring the number of evaluation steps before an episode terminates. In RoboTwin evaluation, a successful episode terminates immediately after the success condition is satisfied, while unsuccessful episodes continue until the evaluation horizon. As shown in Figure~\ref{fig:execution-efficiency}, \method{} reduces the mean episode length from 156.28 to 132.80 steps across all rollouts, a 15.02\% reduction relative to Motus. This aggregate comparison combines successful and failed episodes and is therefore affected by both policy performance and horizon-limited failures. Conditioning on successful rollouts provides a more direct measure of execution efficiency: \method{} reduces the mean length from 79.57 to 74.82 steps, corresponding to a 5.97\% reduction while preserving task completion. This result is consistent with stage guidance helping the policy progress more directly toward intermediate task targets and reducing redundant local corrections.

\section{Conclusion}
We introduced \method{}, which complements the short-horizon visual and action prediction of a WAM with goal-conditioned JEPA guidance for the next stage of task progress. Across 50 RoboTwin tasks, \method{} achieves 91.42\% success in clean environments and 89.08\% under randomization, outperforming Motus across most semantic task families. Component ablations demonstrate the contribution of both stage prediction and its integration into the WAM representation, while execution analysis shows that \method{} completes successful episodes with shorter trajectories. Together, these results demonstrate the effectiveness of stage-level future guidance and the importance of integrating it with local world--action modeling. Future work will extend this direction along two robotics-centered axes: reducing the inference latency of WAMs to enable higher-frequency closed-loop control, and learning multi-scale latent representations that jointly encode coarse-grained task progress and the fine-grained contact, geometry, and motion information required for precise manipulation.
\newpage

\bibliography{main}
\newpage
\appendix

\section{Implementation Details}
\label{sec:app-implementation}

\subsection{Stage-pair construction.}
Stage-pair construction uses frozen V-JEPA2 features to detect salient transitions in each demonstration. For an episode with $T$ decoded frames, we first sample candidate center frames every eight frames and include the terminal frame $T-1$. If this produces more than 32 candidates, we uniformly subsample the ordered list to 32. We denote the resulting candidate centers by $\mathcal C=(c_0,\ldots,c_{M-1})$, where $M\leq32$. A centered 64-frame clip is encoded at every $c_j$; indices outside $[0,T-1]$ are clipped to the nearest endpoint.

For each candidate-list position $j$, we compare positions $j^-=\max(0,j-2)$ and $j^+=\min(M-1,j+2)$. The clip-level score is the Euclidean distance between their mean-pooled V-JEPA2 representations. The token-level score is the maximum Euclidean distance over corresponding token positions. Both score sequences are smoothed with a three-point moving average and independently min--max normalized within the episode; the transition score at $j$ is the larger of the two normalized scores. Greedy temporal non-maximum suppression (NMS) then visits candidates in descending score order. Selecting position $j$ suppresses candidate-list positions $k$ satisfying $|k-j|\leq3$, and selection stops after at most five candidates. Thus, the NMS radius is measured in positions of the retained candidate list, not directly in source-frame units.

The selected center frames, together with frames 0 and $T-1$, form an ordered boundary sequence $\mathcal B=(b_0,\ldots,b_L)$. For every interval $[b_i,b_{i+1})$, each integer frame $t$ is paired with the next boundary $b_{i+1}$. Hence, one interval contributes $b_{i+1}-b_i$ dense training pairs $(t,b_{i+1})$; the selected keyframes define pair boundaries but are not manual semantic-stage annotations. Training converts each pair into a current slice centered at $t$ and a target slice centered at $b_{i+1}$. Both use source-frame offsets $\{-32,-31,\ldots,31\}$, and clipping at an episode endpoint repeats that endpoint frame. These centered slices are an offline demonstration-supervision construction and may contain observations on both sides of their center; closed-loop inference does not have access to target slices or unobserved future frames.

\subsection{Training details.}
Stage I freezes the V-JEPA2 encoder and optimizes the pretrained V-JEPA2 predictor and task-instruction adapter with AdamW. Frozen encoding is performed in FP16, while the learnable modules are trained in FP32 for 1,000 optimizer updates with learning rate $1\times10^{-5}$, weight decay 0.01, and 100 warmup updates. Its per-GPU batch size is 32 on eight NVIDIA A800 80GB GPUs, with no gradient accumulation, giving an effective batch size of $32\times8=256$. The online V-JEPA2 encoding micro-batch of eight clips controls encoder memory use and does not change the optimizer batch size.

Stage II freezes the complete Stage-JEPA branch and trains the local world--action policy and conditioning adapter with AdamW. Let the condition frame have raw demonstration index $t$ (offset 0). The action target has $H_a=16$ samples at indices $t+\{3,6,\ldots,48\}$, and the video-prediction target has $H_v=8$ frames at indices $t+\{6,12,\ldots,48\}$. These offsets count decoded frames in the recorded RoboTwin trajectory: they mean that both targets span through raw frame $t+48$, not that the policy predicts 48 actions. Target video frames have resolution $384\times320$. The video and action objectives both have unit weight. Training uses BF16 for four complete data epochs with learning rate $1\times10^{-6}$, weight decay 0.01, and 200 warmup updates. The per-GPU batch size is four, and gradients are accumulated for eight iterations on eight A800 80GB GPUs, giving an effective batch size of $4\times8\times8=256$. Stage I is trained once with seed 0 for 1,000 optimizer steps, and the final step-1,000 checkpoint is used by Stage II. Stage II is trained once with seed 0 for four epochs, and the final epoch-four checkpoint is used for closed-loop evaluation.

\paragraph{Gate and token sampling.}
V-JEPA2 receives 64-frame $256\times256$ clips. With temporal tubelet size 2 and spatial patch size 16, each clip produces $32\times16\times16=8192$ tokens of dimension 1024. For Stage-JEPA prediction, we uniformly sample 512 context-token positions and 256 target-token positions from the flattened token sequence using rounded linearly spaced indices. The predictor outputs 256 target tokens, which are mean-pooled to form the stage representation used by the WAM conditioning interface. The stage condition is injected through a single global scalar gate $\alpha=0.2\sigma(\beta)$ initialized to 0.02. This scalar is shared across samples, layers, video tokens, and channels; the injected condition vector itself remains sample-specific and is broadcast to all video tokens of that sample.

\begin{table}[t]
\centering
\small
\setlength{\tabcolsep}{3pt}
\begin{tabular}{@{}lp{0.56\columnwidth}r@{}}
\toprule
\textbf{Task} & \textbf{Terminal state} & \textbf{Score} \\
\midrule
\multirow{4}{*}{Cube\_Grasp}
& The cube is not grasped. & 0 \\
& The cube is grasped but remains on the table. & 30 \\
& The cube is lifted but dropped before the end of the rollout. & 70 \\
& The cube is lifted and stably held until the end of the rollout. & 100 \\
\midrule
\multirow{4}{*}{Power\_Strip}
& The power strip is not moved to the target area. & 0 \\
& The left arm moves the power strip to the center position. & 30 \\
& The power strip is stabilized and the right end-effector is aligned with the switches. & 60 \\
& All required switches are successfully pressed. & 100 \\
\bottomrule
\end{tabular}
\caption{Task-progress scoring protocol for the LIFT2 real-robot tasks .}
\label{tab:real-robot-scoring}
\end{table}

\begin{table}[t]
\centering
\small
\setlength{\tabcolsep}{3pt}
\begin{tabular}{@{}lccc@{}}
\toprule
\textbf{Method} & \textbf{Cube\_Grasp} & \textbf{Power\_Strip} & \textbf{Average} \\
\midrule
$\pi_{0.5}$ & 33.50 & 49.50 & 41.50 \\
\method{} & \textbf{45.00} & \textbf{51.00} & \textbf{48.00} \\
\bottomrule
\end{tabular}
\caption{LIFT2 real-robot evaluation. Each entry is the mean task-progress score; higher is better.}
\label{tab:app-real-robot-results}
\end{table}

\begin{table*}[t]
\centering
\small
\setlength{\tabcolsep}{4.5pt}
\begin{tabular}{@{}lrrrrrr@{}}
\toprule
\textbf{Split} 
& \textbf{Episodes} 
& \textbf{Pairs} 
& \textbf{Mean Dist.} 
& \textbf{Median Dist.} 
& \textbf{Cross Boundary} 
& \textbf{Slice Overlap} \\
\midrule
Clean      & 2,482  & 546,152   & 33.90 & 24 & 84.67 & 86.77 \\
Randomized & 24,762 & 5,461,957 & 34.63 & 25 & 82.97 & 86.34 \\
\bottomrule
\end{tabular}
\caption{Diagnostics of the automatically constructed Stage-I supervision. ``Mean Dist.'' and ``Median Dist.'' report the mean and median distance, in frames, from each dense current frame $t$ to its paired next-stage boundary $b_{i+1}$. ``Cross Boundary'' reports the percentage of centered 64-frame current slices that include frames outside the assigned stage interval, and ``Slice Overlap'' reports the percentage of current--target slice pairs that share at least one decoded source frame.}
\label{tab:stage-boundary-diagnostics}
\end{table*}

\subsection{Closed-loop inference.}
At inference, the policy maintains a causal 64-frame observation buffer. The buffer contains only observations that have already been produced by the environment; before 64 observations are available, the earliest observed frame is repeated for padding. Stage boundaries and target slices are not available at inference time. The frozen Stage-JEPA branch predicts the stage condition from the causal observation buffer and task instruction, and the WAM then predicts the finite-horizon visual future and action chunk for closed-loop execution.

\paragraph{Stage-boundary and slice-overlap diagnostics.}
We further diagnose the automatically constructed Stage-I supervision to characterize its temporal scale and the effect of using centered V-JEPA2 input windows. Let $b_i$ and $b_{i+1}$ denote two consecutive stage boundaries. Dense supervision pairs every frame $t\in[b_i,b_{i+1})$ with the next boundary $b_{i+1}$, and the corresponding stage length is $\ell_i=b_{i+1}-b_i$. The diagnostic set covers 100 manifests, including 50 RoboTwin tasks in both clean and randomized splits, and contains 27,244 episodes, 134,285 stage intervals, and 6,008,109 dense pairs after applying the same data-validity checks as the training loader. The inferred stages are relatively short: the mean stage length is 44.74 frames, the median stage length is 34 frames, and 81.72\% of intervals are shorter than the 64-frame V-JEPA2 input window. As a result, centered 64-frame current slices often cross the assigned stage interval or overlap with their paired target slices. Table~\ref{tab:stage-boundary-diagnostics} quantifies this effect at the pair and slice level. This overlap is a property of offline representation-level supervision from demonstrations: Stage-I learns a temporally contextual next-stage representation rather than a strict single-frame extrapolation target.

\begin{table*}[t]
\centering
\begingroup
\small
\setlength{\tabcolsep}{1mm}
\begin{tabular}{@{}l*{10}{r}@{}}
\toprule
\textbf{Simulation Task}
  & \multicolumn{2}{c}{\textbf{GO-1}}
  & \multicolumn{2}{c}{\boldmath$\pi_{0.5}$}
  & \multicolumn{2}{c}{\textbf{X-VLA}}
  & \multicolumn{2}{c}{\textbf{Motus}}
  & \multicolumn{2}{c}{\textbf{\method{}}} \\
\cmidrule(lr){2-3}
\cmidrule(lr){4-5}
\cmidrule(lr){6-7}
\cmidrule(lr){8-9}
\cmidrule(l){10-11}
  & Clean & Rand. & Clean & Rand. & Clean & Rand. & Clean & Rand. & Clean & Rand. \\
\midrule
\emph{Adjust Bottle} & 49\% & 62\% & 79\% & 83\% & \textbf{100\%} & \textbf{99\%} & 89\% & 93\% & 93\% & 94\% \\
\emph{Beat Block Hammer} & 6\% & 10\% & 63\% & 50\% & 92\% & 88\% & 95\% & 88\% & \textbf{97\%} & \textbf{94\%} \\
\emph{Blocks Ranking Rgb} & 7\% & 3\% & 43\% & 35\% & 83\% & 83\% & 99\% & \textbf{97\%} & \textbf{100\%} & 96\% \\
\emph{Blocks Ranking Size} & 2\% & 2\% & 8\% & 14\% & 67\% & \textbf{74\%} & \textbf{75\%} & 63\% & 72\% & 72\% \\
\emph{Click Alarmclock} & 95\% & 90\% & 97\% & 93\% & 99\% & 99\% & \textbf{100\%} & \textbf{100\%} & \textbf{100\%} & \textbf{100\%} \\
\emph{Click Bell} & 98\% & 95\% & 75\% & 76\% & \textbf{100\%} & \textbf{100\%} & \textbf{100\%} & \textbf{100\%} & \textbf{100\%} & \textbf{100\%} \\
\emph{Dump Bin Bigbin} & 57\% & 45\% & 30\% & 42\% & 79\% & 77\% & 95\% & 91\% & \textbf{98\%} & \textbf{95\%} \\
\emph{Grab Roller} & 99\% & 99\% & 90\% & 89\% & \textbf{100\%} & \textbf{100\%} & \textbf{100\%} & \textbf{100\%} & \textbf{100\%} & 99\% \\
\emph{Handover Block} & 9\% & 12\% & 18\% & 19\% & 73\% & 37\% & 86\% & 73\% & \textbf{93\%} & \textbf{87\%} \\
\emph{Handover Mic} & 12\% & 8\% & 28\% & 18\% & 0\% & 0\% & 78\% & 63\% & \textbf{100\%} & \textbf{100\%} \\
\emph{Hanging Mug} & 0\% & 0\% & 3\% & 3\% & 23\% & 27\% & 38\% & \textbf{38\%} & \textbf{40\%} & 30\% \\
\emph{Lift Pot} & 92\% & 92\% & 0\% & 0\% & \textbf{99\%} & \textbf{100\%} & 96\% & 99\% & \textbf{99\%} & 99\% \\
\emph{Move Can Pot} & 16\% & 4\% & 29\% & 27\% & \textbf{89\%} & 86\% & 34\% & 74\% & 85\% & \textbf{89\%} \\
\emph{Move Pillbottle Pad} & 9\% & 11\% & 33\% & 29\% & 73\% & 71\% & 93\% & 96\% & \textbf{98\%} & \textbf{98\%} \\
\emph{Move Playingcard Away} & 37\% & 24\% & 59\% & 67\% & 93\% & \textbf{98\%} & \textbf{100\%} & 96\% & \textbf{100\%} & 96\% \\
\emph{Move Stapler Pad} & 3\% & 4\% & 16\% & 18\% & 78\% & 73\% & 83\% & \textbf{85\%} & \textbf{86\%} & 83\% \\
\emph{Open Laptop} & 65\% & 60\% & 19\% & 35\% & 93\% & \textbf{100\%} & \textbf{95\%} & 91\% & 92\% & 96\% \\
\emph{Open Microwave} & 12\% & 14\% & 35\% & 37\% & 79\% & 71\% & 95\% & 91\% & \textbf{100\%} & \textbf{99\%} \\
\emph{Pick Diverse Bottles} & 61\% & 56\% & 5\% & 3\% & 58\% & 36\% & \textbf{90\%} & \textbf{91\%} & 81\% & 83\% \\
\emph{Pick Dual Bottles} & 81\% & 74\% & 10\% & 6\% & 47\% & 36\% & \textbf{96\%} & \textbf{90\%} & 90\% & 85\% \\
\emph{Place A2b Left} & 33\% & 36\% & 62\% & 60\% & 48\% & 49\% & 88\% & 79\% & \textbf{92\%} & \textbf{88\%} \\
\emph{Place A2b Right} & 31\% & 22\% & 62\% & 57\% & 36\% & 36\% & 91\% & \textbf{87\%} & \textbf{95\%} & 85\% \\
\emph{Place Bread Basket} & 47\% & 52\% & 48\% & 56\% & 81\% & 71\% & \textbf{91\%} & \textbf{94\%} & 89\% & 93\% \\
\emph{Place Bread Skillet} & 2\% & 1\% & 38\% & 46\% & 77\% & 67\% & 86\% & 83\% & \textbf{90\%} & \textbf{88\%} \\
\emph{Place Burger Fries} & 88\% & 92\% & 66\% & 70\% & 94\% & 94\% & \textbf{98\%} & \textbf{98\%} & 97\% & 93\% \\
\emph{Place Can Basket} & 29\% & 37\% & 19\% & 25\% & 49\% & 52\% & 81\% & 76\% & \textbf{85\%} & \textbf{84\%} \\
\emph{Place Cans Plasticbox} & 68\% & 77\% & 40\% & 47\% & 97\% & 98\% & 98\% & 94\% & \textbf{100\%} & \textbf{100\%} \\
\emph{Place Container Plate} & 73\% & 70\% & 71\% & 78\% & 97\% & 95\% & \textbf{98\%} & \textbf{99\%} & \textbf{98\%} & 98\% \\
\emph{Place Dual Shoes} & 6\% & 10\% & 12\% & 7\% & 79\% & 88\% & 93\% & 87\% & \textbf{95\%} & \textbf{96\%} \\
\emph{Place Empty Cup} & 44\% & 39\% & 75\% & 86\% & \textbf{100\%} & 98\% & 99\% & 98\% & 98\% & \textbf{99\%} \\
\emph{Place Fan} & 1\% & 0\% & 25\% & 36\% & 80\% & 75\% & 91\% & \textbf{87\%} & \textbf{94\%} & 86\% \\
\emph{Place Mouse Pad} & 15\% & 10\% & 21\% & 26\% & 70\% & \textbf{70\%} & 66\% & 68\% & \textbf{71\%} & 66\% \\
\emph{Place Object Basket} & 48\% & 49\% & 43\% & 36\% & 44\% & 39\% & \textbf{81\%} & \textbf{87\%} & \textbf{81\%} & 80\% \\
\emph{Place Object Scale} & 26\% & 27\% & 40\% & 49\% & 52\% & 74\% & 88\% & 85\% & \textbf{90\%} & \textbf{93\%} \\
\emph{Place Object Stand} & 56\% & 63\% & 74\% & 65\% & 86\% & 88\% & \textbf{98\%} & 97\% & 97\% & \textbf{100\%} \\
\emph{Place Phone Stand} & 30\% & 37\% & 49\% & 53\% & 88\% & 87\% & 87\% & 86\% & \textbf{90\%} & \textbf{88\%} \\
\emph{Place Shoe} & 15\% & 13\% & 57\% & 61\% & 96\% & 95\% & 99\% & 97\% & \textbf{100\%} & \textbf{99\%} \\
\emph{Press Stapler} & 66\% & 51\% & 80\% & 70\% & 92\% & 98\% & 93\% & 98\% & \textbf{98\%} & \textbf{100\%} \\
\emph{Put Bottles Dustbin} & 7\% & 4\% & 12\% & 9\% & 74\% & 77\% & 81\% & 79\% & \textbf{92\%} & \textbf{86\%} \\
\emph{Put Object Cabinet} & 60\% & 43\% & 24\% & 15\% & 46\% & 48\% & \textbf{88\%} & \textbf{71\%} & 65\% & 54\% \\
\emph{Rotate Qrcode} & 22\% & 9\% & 47\% & 56\% & 34\% & 33\% & 89\% & 73\% & \textbf{91\%} & \textbf{81\%} \\
\emph{Scan Object} & 1\% & 2\% & 42\% & 38\% & 14\% & 36\% & 67\% & \textbf{66\%} & \textbf{85\%} & 52\% \\
\emph{Shake Bottle Horizontally} & 97\% & 92\% & 96\% & \textbf{100\%} & \textbf{100\%} & \textbf{100\%} & \textbf{100\%} & 98\% & \textbf{100\%} & 95\% \\
\emph{Shake Bottle} & 97\% & 93\% & 91\% & \textbf{100\%} & 99\% & \textbf{100\%} & \textbf{100\%} & 97\% & \textbf{100\%} & \textbf{100\%} \\
\emph{Stack Blocks Three} & 1\% & 1\% & 15\% & 16\% & 6\% & 10\% & 91\% & \textbf{95\%} & \textbf{97\%} & \textbf{95\%} \\
\emph{Stack Blocks Two} & 12\% & 22\% & 48\% & 56\% & 92\% & 87\% & \textbf{100\%} & \textbf{98\%} & \textbf{100\%} & \textbf{98\%} \\
\emph{Stack Bowls Three} & 4\% & 7\% & 33\% & 35\% & 76\% & 86\% & 79\% & \textbf{87\%} & \textbf{82\%} & 86\% \\
\emph{Stack Bowls Two} & 51\% & 45\% & 78\% & 66\% & 96\% & 93\% & \textbf{98\%} & \textbf{98\%} & 97\% & 97\% \\
\emph{Stamp Seal} & 19\% & 13\% & 36\% & 23\% & 76\% & 82\% & 93\% & 92\% & \textbf{98\%} & \textbf{98\%} \\
\emph{Turn Switch} & 34\% & 30\% & 5\% & 6\% & 40\% & 61\% & \textbf{84\%} & 78\% & 80\% & \textbf{81\%} \\
\midrule
\textbf{Average (\%)} & 37.80 & 36.24 & 42.98 & 43.84 & 72.80 & 72.84 & 88.66 & 87.02 & \textbf{91.42} & \textbf{89.08} \\
\bottomrule
\end{tabular}
\endgroup
\caption{Appendix: full per-task evaluation on 50 RoboTwin 2.0 simulation tasks under clean and randomized settings. Baseline results are reproduced from Motus~\cite{bi2025motusunifiedlatentaction}; their printed aggregate values are preserved in the final row. Each task--setting cell reports success over 100 closed-loop trials. Bold denotes the highest success rate in each row and setting, including ties.}
\label{tab:robotwin-full}
\end{table*}

\section{More Experimental Results}
\label{sec:app-more-results}

\subsection{Complete RoboTwin task-level results.}

Table~\ref{tab:robotwin-full} reports the complete 50-task RoboTwin results used to compute the semantic-category averages in the main paper. Each entry reports success over 100 closed-loop trials for the corresponding task and setting.

\subsection{Real-world evaluation}

The real-world evaluation is conducted on an in-house dual-arm LIFT2 (ARX R5) platform. We report two tasks compared with $\pi_{0.5}$~\cite{intelligence2025pi_05}. Cube\_Grasp requires the robot to approach a small cube, grasp it, lift it from the table, and keep it stable until the rollout ends. Power\_Strip requires coordinated bimanual manipulation: the left arm moves and stabilizes the power strip, while the right arm aligns with and presses the target switches. Each task is evaluated over 20 trials and we report the average results.

\paragraph{Scoring.}
We score each rollout with a task-progress score from 0 to 100. As shown in Table~\ref{tab:real-robot-scoring}, the score is assigned according to the terminal state of the rollout, and higher scores indicate greater task progress. Safety failures, including dangerous collisions, uncontrolled motions, workspace violations, or emergency stops, receive a score of 0.

\paragraph{Results.}

We trained a separate model for each task, using 50 trajectories for Cube Grasp and 116 trajectories for Power Strip. Table~\ref{tab:app-real-robot-results} reports the mean task-progress score on the two LIFT2 real-robot tasks. We report only average scores for each task and the unweighted average across tasks.

\begin{figure*}[t]
    \centering
    \begin{subfigure}[t]{0.30\linewidth}
        \centering
        \includegraphics[width=\linewidth]{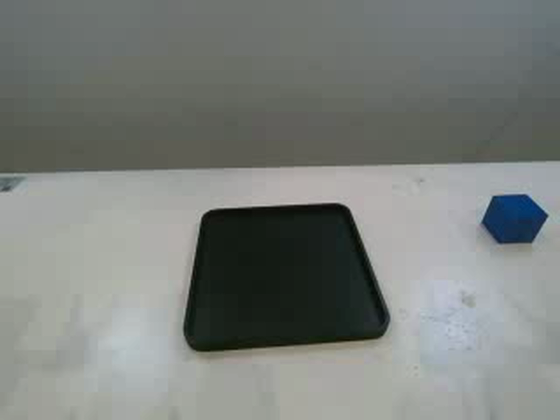}
        \caption{Cube: initial object state.}
        \label{fig:real-robot-cube-setup}
    \end{subfigure}
    \hfill
    \begin{subfigure}[t]{0.30\linewidth}
        \centering
        \includegraphics[width=\linewidth]{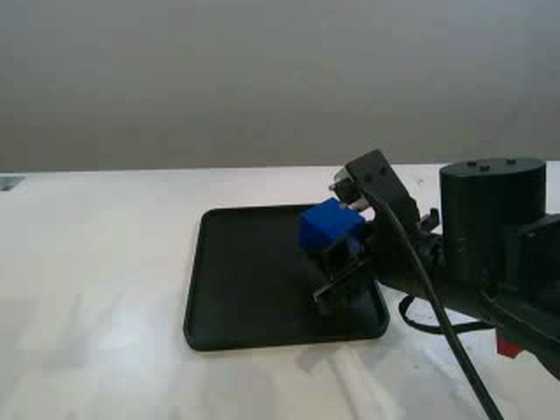}
        \caption{Cube: grasping and lifting.}
        \label{fig:real-robot-cube-interaction}
    \end{subfigure}
    \hfill
    \begin{subfigure}[t]{0.30\linewidth}
        \centering
        \includegraphics[width=\linewidth]{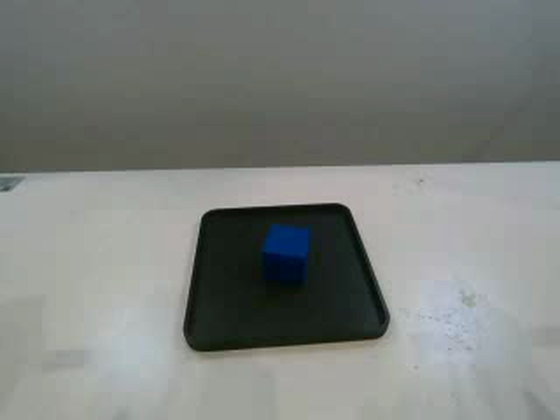}
        \caption{Cube: terminal task state.}
        \label{fig:real-robot-cube-terminal}
    \end{subfigure}

    \begin{subfigure}[t]{0.30\linewidth}
        \centering
        \includegraphics[width=\linewidth]{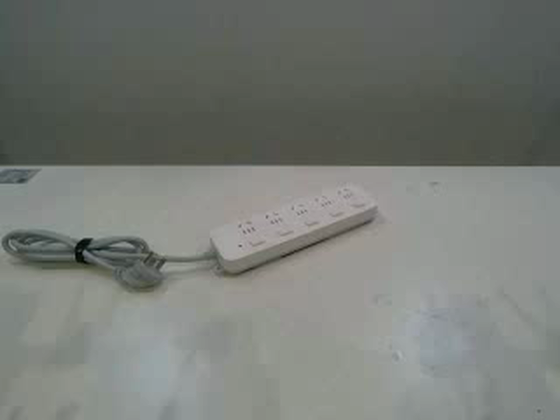}
        \caption{Power: initial object state.}
        \label{fig:real-robot-power-setup}
    \end{subfigure}
    \hfill
    \begin{subfigure}[t]{0.30\linewidth}
        \centering
        \includegraphics[width=\linewidth]{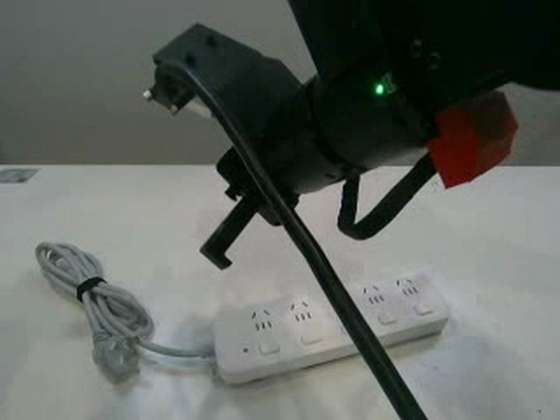}
        \caption{Power: arm--switch interaction.}
        \label{fig:real-robot-power-interaction}
    \end{subfigure}
    \hfill
    \begin{subfigure}[t]{0.30\linewidth}
        \centering
        \includegraphics[width=\linewidth]{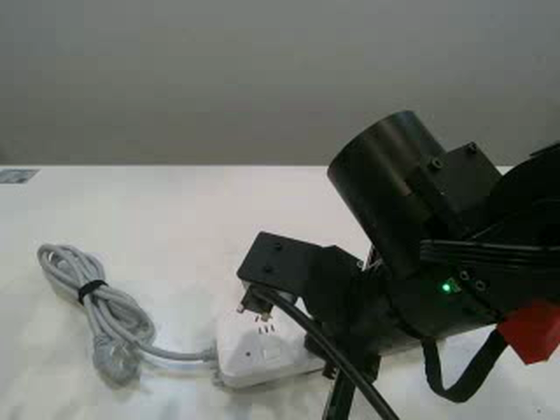}
        \caption{Power: terminal task state.}
        \label{fig:real-robot-power-terminal}
    \end{subfigure}
    \caption{Real-robot task-stage visualization on the LIFT2 platform. The top row shows cube grasping, and the bottom row shows power-strip operation. For each task, the panels show the initial object state, the main interaction stage, and the terminal task state.}
    \label{fig:real-robot-task-stages}
\end{figure*}

\begin{figure*}[t]
    \centering
    \begin{subfigure}[t]{0.30\linewidth}
        \centering
        \includegraphics[width=\linewidth]{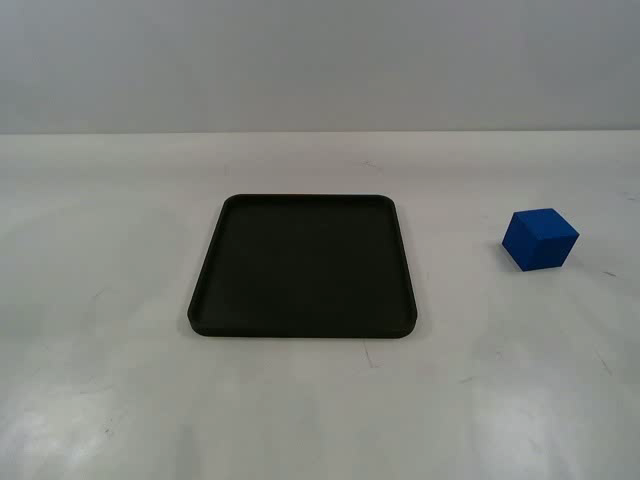}
        \caption{Initial cube state.}
        \label{fig:real-robot-cube-rollout-initial}
    \end{subfigure}
    \hfill
    \begin{subfigure}[t]{0.30\linewidth}
        \centering
        \includegraphics[width=\linewidth]{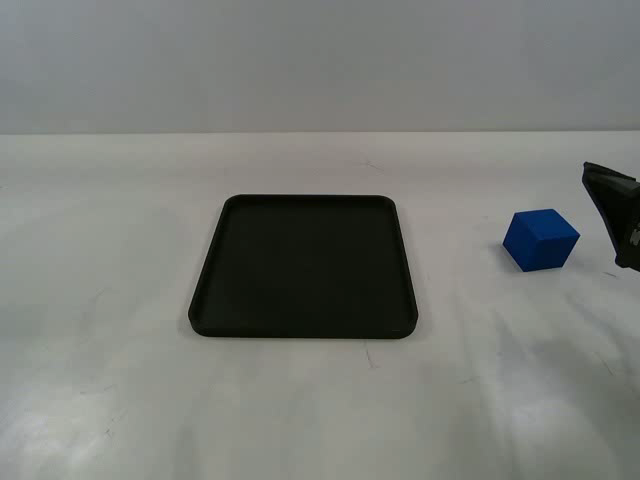}
        \caption{Policy approaches the cube.}
        \label{fig:real-robot-cube-rollout-approach}
    \end{subfigure}
    \hfill
    \begin{subfigure}[t]{0.30\linewidth}
        \centering
        \includegraphics[width=\linewidth]{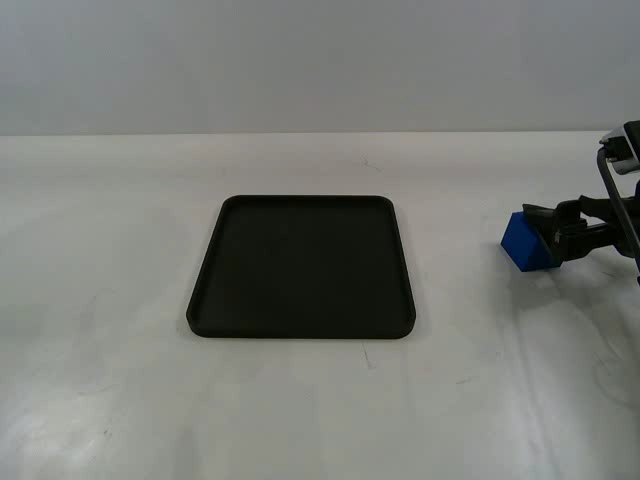}
        \caption{End-effector contacts the cube.}
        \label{fig:real-robot-cube-rollout-contact}
    \end{subfigure}

    \begin{subfigure}[t]{0.30\linewidth}
        \centering
        \includegraphics[width=\linewidth]{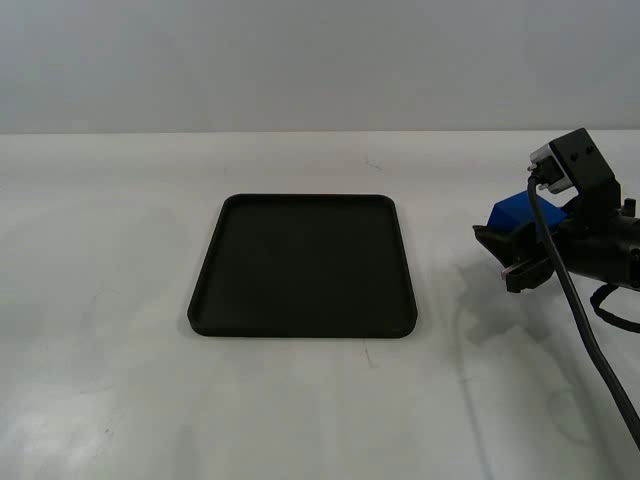}
        \caption{Cube is lifted.}
        \label{fig:real-robot-cube-rollout-lift}
    \end{subfigure}
    \hfill
    \begin{subfigure}[t]{0.30\linewidth}
        \centering
        \includegraphics[width=\linewidth]{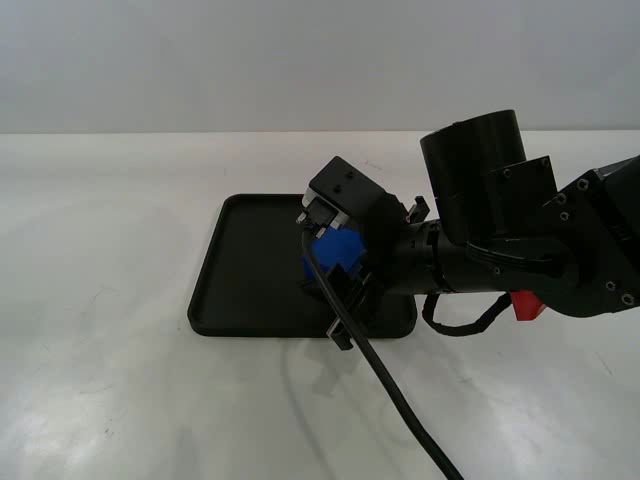}
        \caption{Cube is moved over the tray.}
        \label{fig:real-robot-cube-rollout-transfer}
    \end{subfigure}
    \hfill
    \begin{subfigure}[t]{0.30\linewidth}
        \centering
        \includegraphics[width=\linewidth]{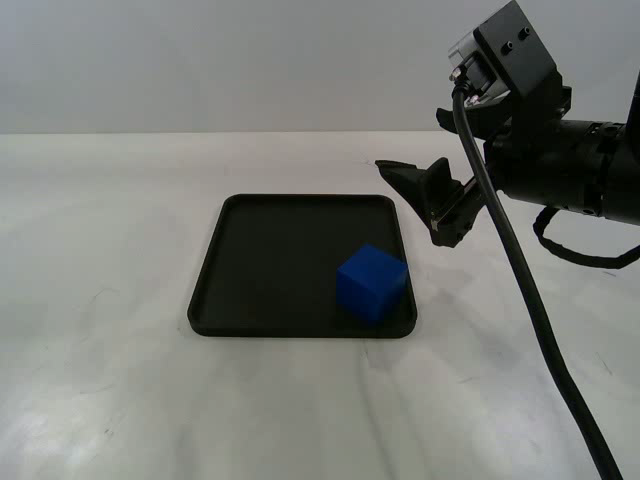}
        \caption{Terminal task state.}
        \label{fig:real-robot-cube-rollout-terminal}
    \end{subfigure}
    \caption{Representative closed-loop \method{} rollout on the LIFT2 cube-grasping task. The panels show the policy progressing from the initial state, through approach, contact, lifting, and transfer, to the terminal task state.}
    \label{fig:real-robot-cube-rollout}
\end{figure*}

\begin{figure*}[t]
    \centering
    \begin{subfigure}[t]{0.30\linewidth}
        \centering
        \includegraphics[width=\linewidth]{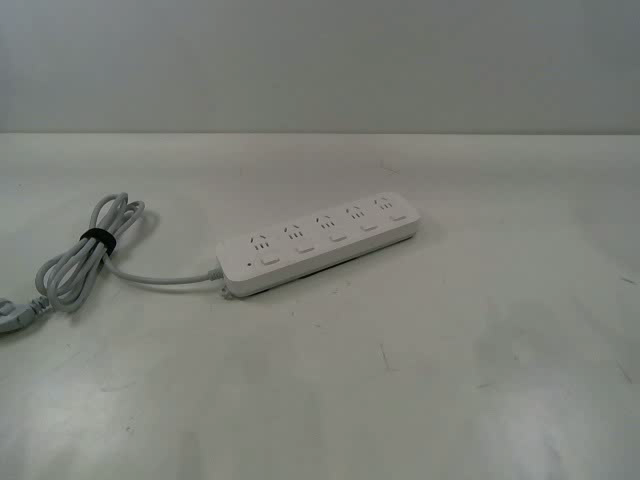}
        \caption{Initial power-strip state.}
        \label{fig:real-robot-power-rollout-initial}
    \end{subfigure}
    \hfill
    \begin{subfigure}[t]{0.30\linewidth}
        \centering
        \includegraphics[width=\linewidth]{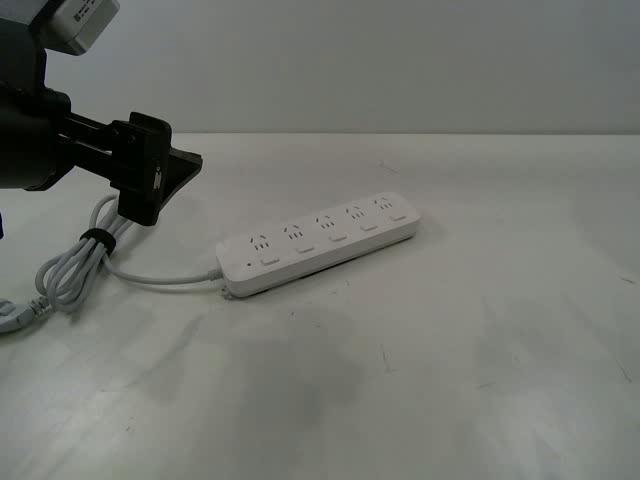}
        \caption{Policy approaches the strip.}
        \label{fig:real-robot-power-rollout-approach}
    \end{subfigure}
    \hfill
    \begin{subfigure}[t]{0.30\linewidth}
        \centering
        \includegraphics[width=\linewidth]{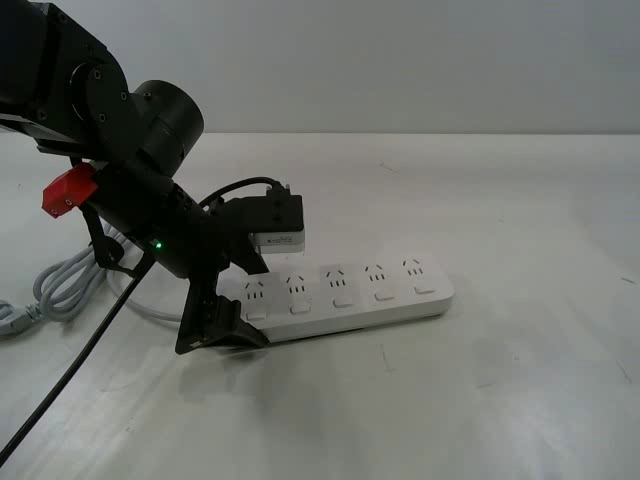}
        \caption{End-effector contacts the strip.}
        \label{fig:real-robot-power-rollout-contact}
    \end{subfigure}

    \begin{subfigure}[t]{0.30\linewidth}
        \centering
        \includegraphics[width=\linewidth]{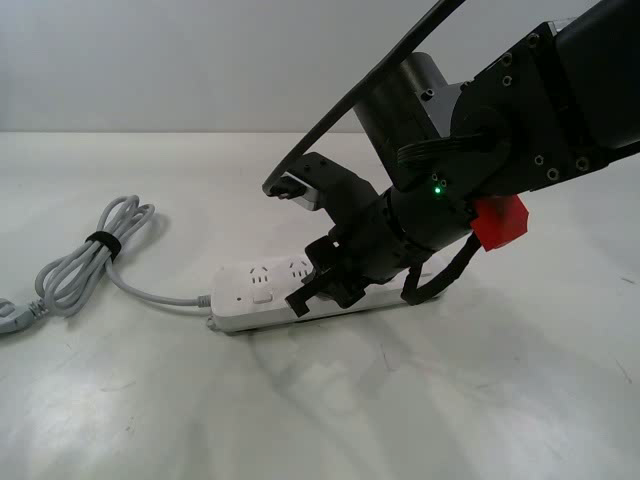}
        \caption{Policy aligns with switches.}
        \label{fig:real-robot-power-rollout-align}
    \end{subfigure}
    \hfill
    \begin{subfigure}[t]{0.30\linewidth}
        \centering
        \includegraphics[width=\linewidth]{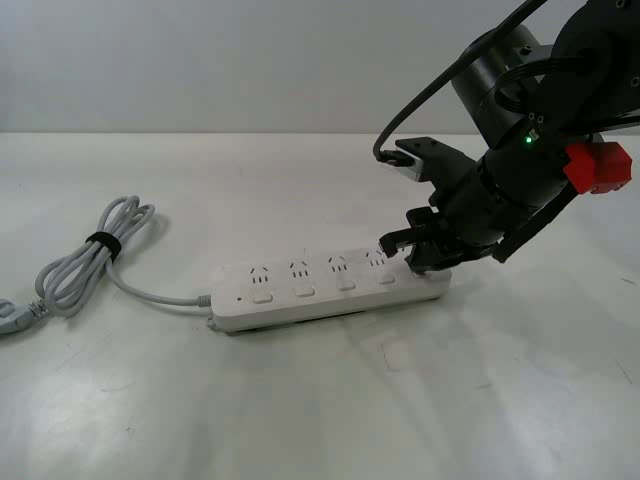}
        \caption{Switch-pressing interaction.}
        \label{fig:real-robot-power-rollout-press}
    \end{subfigure}
    \hfill
    \begin{subfigure}[t]{0.30\linewidth}
        \centering
        \includegraphics[width=\linewidth]{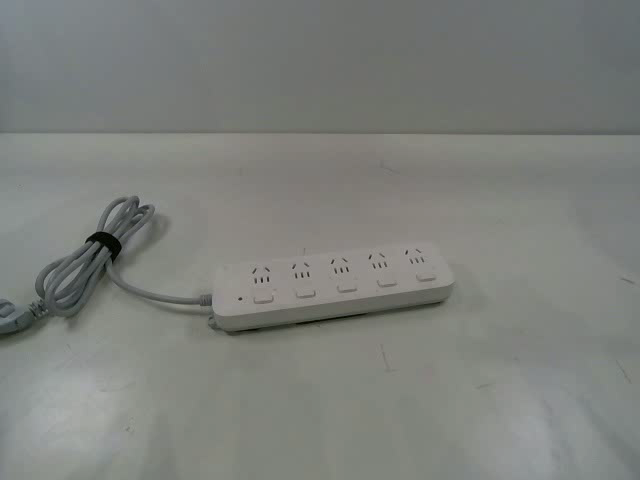}
        \caption{Terminal task state.}
        \label{fig:real-robot-power-rollout-terminal}
    \end{subfigure}
    \caption{Representative closed-loop \method{} rollout on the LIFT2 power-strip task. The panels show the policy progressing from the initial state, through approach, contact, alignment, and switch-pressing interaction, to the terminal task state.}
    \label{fig:real-robot-power-rollout}
\end{figure*}

\paragraph{Real-robot visualizations.}
Figure~\ref{fig:real-robot-task-stages} visualizes the ordered stages of the real-robot tasks on the LIFT2 platform. Figures~\ref{fig:real-robot-cube-rollout} and~\ref{fig:real-robot-power-rollout} show representative closed-loop \method{} rollouts on the cube-grasping and power-strip tasks, respectively.

\end{document}